# File mapping Rule-based DBMS and Natural Language Processing

Vjacheslav M. Novikov
hbm@mail.radiant.ru
VDO RSC Energia, Samara, Russian

**Abstract**

This paper describes the system of storage, extract and processing of information structured similarly to the natural language. For recursive inference the system uses the rules having the same representation, as the data. The environment of storage of information is provided with the File Mapping (SHM) mechanism of operating system. In the paper the main principles of construction of dynamic data structure and language for record of the inference rules are stated; the features of available implementation are considered and the description of the application realizing semantic information retrieval on the natural language is given.

## 1. Introduction

The algorithms of information extract and its processing directly depend on structure, in which the information is represented. And the structural solutions can considerably simplify the algorithmic ones, enabling to look at the problem from another point of view.

The modern systems of natural language processing are created, as a rule, on the basis of text data, represented as records of relational databases. The algorithms of processing are realized frequently by Prolog-similar means. Complexities of interaction of the programming languages, based on recursive inference of rules, and relational databases were earlier considered despite of external analogy of logical interpretation (in more detail in section 3). Besides the tabular form of representation is not always adequate to internal structure of the information processed. The last statement is correct in particular in relation to the traditional and most popular way of storage of the information, in the form of the natural language texts.

In this paper the concept of storage, extract and processing of information structured similarly to constructions of the natural language, with usage of the mechanism of shared segments of memory (SHM) for operating system Unix or the File Mapping mechanism for operating system Windows is stated. The concept is realized in SHM_db DBMS. Further the description of the language used for record of inference rules is given and the way of automatic generation of rules in learning process is considered.

## 2. Structure and rules of the natural language

Let's look at the problem of structuring of the textual information from the point of view of training and, as Turing offered [9], training the child, initially having no preliminary information on structure, constructions and rules of the natural language. Let's accept also, that the training starts with the written language, first excepting the colloquial one.

The selection of *characters*, *words* and *sentences* in the text information does not raise doubt. The obtaining of skills of their differentiation remains behind the frames of the given research. If to consider words and sentences as structural units of more high levels in comparison with characters, each of structural units on its level can participate in derivation of the definite sequences: combination of characters, combination of words etc. The unit of more high level is concerned with some sequence of structural units of the corresponding lower level.

Besides, in the natural language structure it is possible to distinguish other constructions as well. For example, let's represent sentence - question - answer

*Tom played fair. Did Tom play fair ? Tom played fair.*

as

*((Tom played fair .)( Did Tom play fair ?))->( Tom played fair .);*

In parentheses we shall designate the selected sequences of some structural units. (All entered hereinafter denotations are determined by the described below language of record of the inference rules or, to be more exact, by the features of its realization, and it is possible for them not to coincide with the denotations accepted in logic programming or formal Backus-Naur form.) Let's suppose, that the construction containing the sign **->**, defines a *rule* or *inference rule*. The inference rules will form a new higher structural level in relation to the sentences.

Now we shall assume, that in learning process, which purpose is the semantic structuring of the source texts, two rules are showed as samples:

*((Tom played fair .)( Did Tom play fair ?))->( Tom played fair .);*
*((Bill played fair .)( Did Bill play fair ?))->( Bill played fair .);*

Let's compare appropriate sentences of rules. In case the rules differ only by the sentences containing an identical amount of terms and the differences concern only one of these terms, in this and only this case these rules can be replaced by one, composed from the schemes of the sentences. It can look as

$$(( \begin{bmatrix} Tom \\ Bill \end{bmatrix} \text{ played fair .}) (\text{Did Tom play fair ?})) \rightarrow ( \begin{bmatrix} Tom \\ Bill \end{bmatrix} \text{ played fair .}); \qquad (1)$$

Here the expression in square brackets $\begin{bmatrix} Tom \\ Bill \end{bmatrix}$ plays a role of the disjunction.

Let's consider a more complex case. We assume that the rule (1) already exists and in learning process the next rule is also showed

*((Tom spoke fair .)( Did Tom speak fair ?))->( Tom spoke fair .);*     (2)

For generalization (1) and (2) there can be a construction

$$(( \begin{bmatrix} Tom \\ Bill \end{bmatrix} \begin{bmatrix} played \\ spoke \end{bmatrix} \text{ fair .}) (\text{Did} \begin{bmatrix} Tom \\ Bill \end{bmatrix} \begin{bmatrix} play \\ speak \end{bmatrix} \text{ fair ?})) \rightarrow ( \begin{bmatrix} Tom \\ Bill \end{bmatrix} \begin{bmatrix} played \\ spoke \end{bmatrix} \text{ fair .}) \quad (3)$$

$$|<( \begin{bmatrix} play \\ speak \end{bmatrix} \begin{bmatrix} played \\ spoke \end{bmatrix} ) \begin{bmatrix} (play \ played) \\ (speak \ spoke) \end{bmatrix} >;$$

The sign | is followed by the logical expression specifying conditions, and the expression in angular brackets < > plays the role of conjunction.

The rule (3) includes not only three initial constructions presented in the course of training, but also such as

*((Bill spoke fair .)( Did Bill speak fair ?))->( Bill spoke fair .);*

Generally speaking, some of the constructions represented by the rule can be absent in the natural language, despite of syntactical regularity.

The analysis mentioned above does not claim the linguistic entirety; it only characterizes the initial premise of a structural data representation of the system described.

## 3. To the history of the question

A survey paper by Lashmanov A.V. "Deductive databases: language aspect", published in journal *Databases and hypertext systems* number 9'91 [5] can be considered as the starting point of the research described. The paper was written on the basis of the publications in such issues as *Lecture Notes in Computer Science*, *Journal of Logic Programming*, *The Computer Journal*, *SIGMOD Record*, *Technology and Science of Informatics*, *Information Systems*, *Data and Knowledge Engineering*, *Knowledge Bases Systems* and others. The analysis of state and perspectives of the logical programming languages development in relation to database management systems was given in this paper.

It was marked that the development of logical (deductive) languages of databases goes in two directions. First - in the direction of modification and/or creation of Prolog-similar languages (on the basis of Horn formulas), including the mechanism of inquiries to the database. Second - in the direction from relational languages to deductive databases by creation of the specialized logical programming languages, which can be described as the SQL extension. It was emphasized that despite of rather long history of development, accompanied by high research activity, only some of them succeeded in commercial implementation based in general on Prolog modifications. The tendency marked is likely to be saved until now. Despite of logical relationship of the deductive programming languages and relational databases [6], the interfaces of access to relational bases from the languages of logic programming are not applied widely now as well.

Speaking about natural language processing (NLP), it is necessary to remark that certain internal generality of problems in implementation of NLP-systems with the problems of interaction between the logical programming languages and DBMS is observed. The analysis of publications, accessible from the Internet at least, shows that though there is a definite success of the information extract systems on inquiries in the natural language, the solution, equally suitable for the different natural languages, is not available till now. The scheme of such systems functioning remains, as a rule, the same that is, *the inquiry on the natural language - parser - translation in SQL - analysis of the extracted information*. Quite often parser is being built on the basis of Prolog-similar resources. (See, for example, [1,3,8,10].) In this scheme the parser and fulfillment of inquiries to the database are logically and structurally divided. At the same time the attempts of implementation with usage of relational or Adabas-similar DBMS of the data constructions, described in the previous section, in the combination with effective extraction of information and recursive disclosure of rules have not reached any success.

## 4. File Mapping mechanism

As compared to the approaches described in A.V. Lashmanov's paper, the offered DBMS is founded on data model essentially differed from the traditional one. One of the principal components which permitted to realize the main ideas of the model and gave the first conditional title of DBMS - SHM_db, was the mechanism of shared segments of memory (SHM-segments) of operating system Unix.

The shared segments, having appeared in operating system Unix as a tool of inter-processing connection, were supplemented later by the mechanism of their saving. In Windows NT, Windows 95 the similar operation is fulfilled by API's File Mapping functions.

The mechanism described realizes the idea of non-volatile virtual storage with the characteristics of access, being the best, as a rule, in comparison with access to the file system, that allows to refuse completely from the input-output system calls at programming of DBMS kernel functions. Besides, the process connecting a shared segment, allocates it in the limits of 4 Gb range of addresses for Unix or 2 Gb range for Windows NT. When programming on C the addresses can be stipulated and fixed beforehand. In SHM_db the memory address simultaneously fulfills the function of the relation code or the type of the array. So, for example, in implementation for Unix the relations of the database are allocated always on the negative semiaxis of addresses and the references to the positive one, herewith the relations of more high levels having smaller addresses.

The similar system solutions have allowed:
- firstly, to obtain more high-speed load modules realizing basic functions, at the expense of primary usage of the computer instructions with direct addressing instead of indirect;
- secondly, to use data model, which seem prodigal from the point of view of size, that is, for example, a non-byte representation of characters as relations of 0-th level of the database.

## 5. Some main concepts and definitions

In spite of the fact that at the description of algorithms of DBMS operations, such concepts as "relation", "the scheme of storage of the relation" etc. are widely used, they essentially differ from the concepts of the generally accepted relational data model.

*Relation* is any aggregate of the relations or elementary relation of the same level. Each relation has the pointer on the sequence (which can be empty) of direct references. Each non-elementary relation is being characterized by the inverse references and code. The relations, depending on the level, are distinguished according to the type.

*Level of the relation*. Four levels of relations are identified.
0-th level - characters, elementary numbers, combinations of characters and their relations, combinations of elementary numbers. The elementary relation of 0-th level defines the character code or elementary number. Elementary numbers are hexadecimal numbers from 0x00 up to 0xFF. In the database the numbers described on C as *short*, *integer*, *float* and *double* can be used. All numbers, except for elementary, are represented by sequences of elementary numbers. The order of creation of number sequences is reverse - at first rightmost digits, then left. The non-significant zero in the sequences of elementary numbers and trailing blanks in the sequences of characters, as a rule, are not mapped in the database.

1-st level is words, elementary relations of the 1-st level, combinations of words, elementary relations of the 1-st level and their relations. *The word* is a relation having the inverse reference to the relation of the 0-th level. The elementary relation of the 1-st level is a relation defining an arbitrary sequence of characters (for non-descriptors or arbitrary bit maps), or relation defining the address of the subroutine/function, or empty relation of the 1-st level.

2-nd level is sentences, aggregates of sentences. *The sentence* is a relation having the inverse reference to the relation of the 1-st level.

3-rd level is rules, sequences (or *files*) of rules. *The rule* is a relation having the inverse reference to the relation of the 2-nd level.

*References*. The direct and inverse references are distinguished. The direct references define a set of relations, the given relation being as the constituent. The inverse references define a set of relations, of which the given relation consists.

*Code of the relation*. Except for the elementary relations the codes can be the following ones:
0 - conjunction, in which the order of the relations is important (the term "sequence" is also applied);
1 - the conjunction, in which the order of the relations is indifferent;
2 - the disjunction, in which the order of the relations is indifferent;
3 - the disjunction of sequences of the relations (the term "list" is also applied).

When recording the language sentences the relations with the code 0 are taken in parentheses **( )**, the relations with the code 1 - in angular brackets **< >**, the relations with the code 2 - in square brackets **[ ]** and the relations with the code 3 - in braces **{ }** respectively.

*Paradigm* is the relation with code 2, containing the special relation unambiguously defining paradigm (for the 1-st level a role of special relations is played by variables, for the 2-nd level - by schemes), and the set of relations representing the saved data (the last can be absent). In spite of the fact that a paradigm from the point of view of the search operations logic is perceived as the disjunction, the order of creation of the inverse references can differ from the usual (for the relations with code 2) chronological order on the increase of addresses. If the word defining the paradigm is ended with sign **+,** the inverse references are created in ascending order of paradigm values, if it is ended with sign **-,** the references are created in decreasing order. The sign ` defines the inverse chronological sorting. In other cases the components of the paradigm are written in direct chronological sequence.

*Type of the relation*.
On the 0-th level all relations except for the names of variables have the type 0. A special elementary relation defining the combination of characters as the name of a variable has type 1, and the combination of characters, including this special relation, has type 3.

On the 1-st level variables, global variables and constants are distinguished. The variable (or the current variable) is a paradigm, containing a word with type 3, starting with a small character; it is represented by the relation with type 1. The global variable is a paradigm, containing a word with type 3, starting with a large character, and is represented by the relation with type 2. The constants and constant expressions are referred to type 0. And the relations including variables have type 3.

The sentences on the 2-nd level can be *executable* (code 2), i.e., calling the control transfer to some subroutine/function, or *non-executable* (code 1).

On the 3-rd level *direct* and *inverse* rules are distinguished. The inverse rules are referred to type 1, the direct rules - to type 2. The rules in which the right part is absent are used for the definition of input data. Such rules are referred to type 1. The rules, in which there is no left part, are intended for control transfer immediately from the main program and are referred to type 2.

The types of the relations are stated by the system and are not analyzed by the programmer directly.

*Scheme* (scheme of storage). This concept, applicable only to the relations of the 2-nd level, means a sentence containing variables (both current, and global).

*File*. With reference to SHM_db this is the relation of the 2-nd level containing both the scheme and its corresponding saved data. In other words, it is possible to say that the file is a non-executable paradigm of the 2-nd level.

*File of rules* is the relation of the 3-rd level, containing a set of rules mapping input data of the rules representation program and a special relation defining the time of the last modification of these data and their location.

## 6. The language for record of inference rules

In spite of the fact that the work was primary aimed at programming on C language, the language for the record of inference rules is not the extension C or C++ (as, for example, R++ of AT&T Bell Laboratories [7]). The language has arisen as a means of compact representation of the relations of 4-level dynamic structure, its syntax being non-separable from the main SHM_db concepts and in many respects is defined by them.

The following agreements are accepted for the record of rules.

Here the word is a sequence of characters, digits and characters limited by separators. As separators, the codes of the beginning and extremity of a line, blanks, single characters !"(),/:;=<>?\[]|{} are used (point and sign _ are not separators). The sequences of characters taken in unary quotes ' ' are also considered as words. In double quotes the sequences of words are specified taking into account the separators. All words enclosed in quotes are considered as constants.

The sign # has a special assignment. This sign always precedes the names of subroutines/ functions. The translator of rules in the relation defining the subroutine/function allocates the place for the linkage address. If the first character of the name is large - subroutine/ function is considered as a standard one.

The sign $ in the first position defines the name of the non-executable relation.

Any word not enclosed in quotes or which is not special, is interpreted by translator of rules as the name of a variable, being global if the word starts with a large character, otherwise current.

Special sequences /* */ identify the comments, and == != := += -= >= <= set the relations causing calls to standard subroutines/functions.

For any word in quotes, double or unary, the attempt is made to present it as a number (format is defined by external representation by rules accepted in C), at failure the word is considered a character one.

The separators, as well as relations, are grouped as follows:

**;** is a separator of the 3-rd level, i.e. any rule should be terminated in a semicolon.

**->, |** are the separators of the 2-nd level, they select sentences and groups of sentences, besides **->** separates the left part of the rule from the right, and the sign | separates the conditions.

The level of the relations selected with brackets **( ) < > [ ] { }**, depends on the context. The remaining separators are considered as separators of the 1-st level.

The brackets always should be paired. The parentheses **( )** define the relations having the code 0, the angular brackets **< >** - code 1, the square brackets **[ ]** - code 2 and the braces **{ }** - code 3 respectively.

The executable relations of level 2 always look like:

$$\#xxx \ (z1 \ z2 \ ... \ zN),$$

where **xxx** is the name of the subroutine/function, and **(z1 z2 ... zN)** is the relation with code 0 of the arbitrary number of the 1-st level relations (parameters of the subroutine/ function).

The non-executable relations of the 2-nd level look like:

$$\$xxx \ (z1 \ z2 \ ... \ zN).$$

As compared to the executable relations, they do not call control transfer to any subroutine/ function, but initiate recursive search of derivative rules as the executable relations do. (In the mode of processing of semantic rules the role of non-executable relations is played by sentences and schemes of natural language sentences, but in the language of record of the inference rules the similar constructions have no reflection.)

The rule in a general view is written as:

$$A \text{ -> } B \ | \ C;$$

Here *A* is the left part of the rule, *B* is the right part, *C* is the conditions of transition from *A* to *B*. Sometimes some parts can be absent. The expression:

$$A \text{ ->;}$$

sets the data defined immediately in the file of rules. And the expression:

$$\text{-> } B \ | \ C;$$

means the transition to *B* at execution of conditions *C* immediately from the main program. The unconditional transition from *A* to *B* and unconditional control transfer from the main program respectively is possible also:

$$A \text{ -> } B;$$

$$\text{-> } B;$$

*A* in a general view can be represented as the sequence

$$A1, A2, A3,$$

where  *A1* is the initial sentence, the scheme of sentence A1 or list {A1};
      *A2* is scheme A2, defining some file, or list {A2};
      *A3* is a logical function defining the conditions of transition.

The relation *A1, A2, A3* represents an inverse rule. If *A* has more than one term, the expression is being treated as the check on existence (or search) of data of file A2, which variables accept values corresponding to the relation A1. The search can be completed immediately after finding the first variant, satisfying the criteria of the data search, or be fulfilled cyclically for all possible values. In the latter case A2 is enclosed in curly brackets. At transition without check on existence A2 and A3 are absent.

*B* in a general view can be represented as a sequence

$$B1, B2, ..., Bn.$$

Here each of the constituent relations of level 2 is a sentence, either scheme, or list containing the unique scheme. The list {Bi} in this case is an outcome of exhaustive search of all possible values of uncertain variables or their combinations for the non-executable relations; for the executable relations it means the recurrence of operation.

As a result of executable relation the control is being transmitted sequentially to expressions B1, B2, etc. Normal return code is code 1. The execution of the sequence can be stopped, if one of Bi is ended with a return code 0 or -1. -1 is treated as interruption. For the executable relations the interruptions of the derivative relations are analyses by the executed unit itself.  For non-executable lists {Bi} the completion of one of the components with code -1 results in leaving the list with the same return code.

Each of Bi relations can initiate the process of search of rules, in which A1 is the equivalent to Bi. And, for cutting the review of possible variants of prolongation, the search is started only in the case if Bi defines the non-executable relation or accesses to the subroutine/function completing by the code >= 0410 (initially 0411-0427 were treated as function keys codes of the Unix-terminal).

*C* represents the logical expression

$$L (C1,C2,...,Cn),$$

in which Ci is a sentence or a scheme, and the logic operations AND, OR are set by brackets:
   AND - ( ) or < >,
   OR - [ ] or { },
and in the case of usage of brackets **( )** or **{ }** the procedure for conditions testing is being determined by the order of record of sentences. The negation NOT is being set by the function of the 2-nd level **!**  or  **#Not**.

## 7. Transmission of values of variables

At the relations execution the transmission of values of variables to the following relation is one of the key moments defining the form of algorithm record with the help of the language described. From one relation to another the values of variables are being transmitted as tables of variables. The programs of the DBMS kernel for each of the processes generate tables of two types: the table of global variables TBLVAR and the tables of current variables TBLVar. TBLVAR is always unique - it allows transmitting values of variables from relation to relation irrespective of their enclosure degree. The tables of current variables TBLVar are generated for each executable relation separately.

The modification of the table of global variables or creation of a derived table of current variables happens during the relation execution. For global variables the changes are brought in the same table, further the updated variant is to be used, for current variables the changes have concern only the tables generated (or derived).

Depending on the location of an executable relation in the relation specifying the rule, there are two ways of filling in the tables derived:

1-st, when the relation is the constituent of sequence Bi; the derived tables are generated only on the basis of the parent table irrespective of outcomes of execution of other relations Bi.

2-nd, when the relation is the constituent of sequence Ci; in this case the derived table is unique and coincides with the initial table. All changes, which have been brought in the table at execution of other Ci relations, influence its current state.

The system uses one special variable with the fixed name **key**. The programmer can work with it as with a usual variable without inclusion into the argument list of the subroutine/function. The **key** value assignment happens always after the call to the subroutine defined by the relation from sequence Bi and ended by the return code > 0410.

*Note*. During the search the variables are stated according to the search relation A1 and are being updated by values, which they accept in file A2.

## 8. Subroutines / functions

Every 1-st term of sentences written as **#xxx**, is interpreted as a call to the subroutine (function) with a name **xxx**. If the name of the subroutine starts with a large character, it is considered global (or standard) and is being searched among functions executing by the system kernel.

Not going into details of subroutines/functions description, we shall mark, that some of them are functions of two arguments (#Belong, #Dec, #Eq, #Fix, #Ge, #Grtdat, #Inc, #Le, #Ltldat, #Move, #Ne, #Part, #Spawn), others have one argument (#Date, #List, #SystemR, #Time, #Tstdat), there are subroutines/functions not having arguments (#Break, #Delete, #Exit, #Not, #Save).

Such subroutines/functions as #Delete, #Not, #Save have no arguments too, but play the role of functions of the 2-nd level. Their operation spreads to a sentence, standing in sequences Bi or Ci immediately after these functions. These functions are represented respectively as:

#Delete:
! or #Not:
 #Save:

Functions #Dec, #Eq, #Fix, #Ge, #Inc, #Le, #Ne besides the standard form  #xxx (z1 z2) can also be written also respectively:

$$\begin{aligned}
\#Dec\ (z1\ z2)\ &\text{as}\quad (z1\ -=\ z2)\\
\#Eq\ (z1\ z2)\ &\text{as}\quad (z1\ ==\ z2)\\
\#Fix\ (z1\ z2)\ &\text{as}\quad (z1\ :=\ z2)\\
\#Ge\ (z1\ z2)\ &\text{as}\quad (z1\ >=\ z2)\\
\#Inc\ (z1\ z2)\ &\text{as}\quad (z1\ +=\ z2)\\
\#Le\ (z1\ z2)\ &\text{as}\quad (z1\ <=\ z2)\\
\#Ne\ (z1\ z2)\ &\text{as}\quad (z1\ !=\ z2)
\end{aligned}$$

The return codes of subroutines/functions accept one of three values: 1, 0 and -1, and 1 is treated as an execution of a logical condition, 0 and -1 - as a failure, -1 is considered besides as an interrupt return code.

## 9. The example of application using SHM_db as DBMS

We shall show the characteristic features of the interface and implementation of database inquiries with usage of the language of inference rules on the following example.

We assume that it is required to organize a record in a text database, using as a key the attribute "number of article", and sampling given from base on a key value or in ascending order of attributes values, if the key is not given. We shall limit the external interface to the following windows (Fig.1):

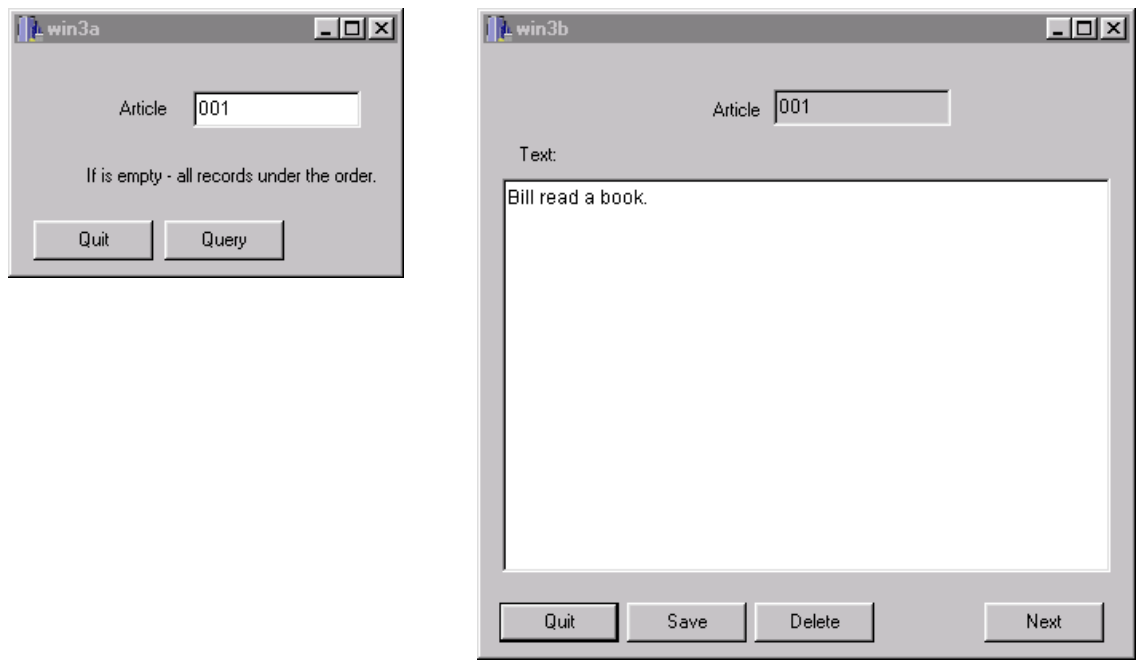

Figure 1.

The algorithm given below has a demonstration character and the optimality of solution was not pursued.

```
param = (art+ {s});
T = (param);
QUERY = 0413;
SAVE = 0423;
DELETE = 0424;

                    /*** Writing in file "Text" ***/

-> (Flag :=0), #win3a(Art), (Flag ==0), $query3b()
                                        | (key == QUERY),(Art :=' ');

#win3a(Art) -> $query3a(art+) | (Art !=' '), (art+ :=Art);

#win3a(Art)  -> { $query3a(art+),} | (Art ==' ');

$query3a(art+), T  -> (Flag :=1), (Old := param), #win3b(param);

$query3b() -> (Old :=" "), #win3b(param) | (Art !=' '),(art+ :=Art);

#win3b(param) -> #Save: T, (Old := param) | (key == SAVE),
                                                        (Old ==" ");

#win3b(param) -> $query3d(), $query3c(param)
                                | (key == SAVE), Old !=" "), (param != Old);

$query3d() -> #Delete: T | (param := Old);

$query3c(param) -> #Save: T, (Old := param);

#win3b(param) -> #Delete: T, #Break() | (key == DELETE);
```

One of the major features of algorithm representation with usage of the language of inference rules is that there is no necessity of the external description of data structure for base. The first five sentences of the given text set substitutions and are introduced only with the purpose of the greater readability of algorithm, and substitution `T = (param);` underlines that the structure of the scheme of storage is identical to the structure of parameters of the window `win3b`.

Setting the system variable `key` should be provided by the programmer at implementation of event processing functions on pressing corresponding keys with the usage of resources of SHM_db program interface. (The creation of the interface units easily gives in to automation. In variant for Unix V.3 the possibility of creation of units with the usage of extended text editor was realized.)

In the given text the first rule states that immediately from the head program, at pressing the key `QUERY` with simultaneous assignment of value `' '`(empty) for global variable `Art`, the following sentences become active sequentially:

`(Flag :=0)` - reset of global variable `Flag` in 0;

`#win3a(Art)` - control transfer to the unit (subroutine/function), realizing the program interface with the window (form) `win3a`;

`(Flag ==0)` - check of `Flag` on 0;

`query3b()` - non-executable sentence generating transition to derivative rules.

The 2-nd rule defines the transition from the unit of window interface `win3a` to non-executable sentence `$query3a(art+)`. The condition of transition is nonblank value `Art`, which is stated at fulfillment of unit `win3a`. Simultaneously this value is assigned to the local variable `art+` (presence of the sign + means that the inverse references of appropriate paradigm are ordered on increasing of values of variable).

The most essential for understanding of logic of operation SHM_db are the 3-rd and the 4-th rules.

The construction `{$query3a (art+),}` means initiation of transition to derivative rules sequentially for all values `art+`.

`query3a(art+), T` is the inverse rule realizing the search of sentences on the value of variable `art+` according to the scheme of storage `T` (in view of substitutions `(art+ {s})` ).

It is necessary to remark, that the field *Text* is described as `{s}`, that is the list of values of `s`(that in the given example is not necessary, as the search on the contents is absent).

The functions of saving and deleting are carried out with the help of functions of the 2-nd level and are being written accordingly as `#Save: T` and `#Delete: T` .

The standard function `#Break()` returns the code -1 to unit `win3b` in case the deleting is produced.

The file containing rules in text format, by resources SHM_db is being analyzed and mapped as the relations of 4-th level's dynamic structure. It is impossible to treat this process, perhaps, as interpretation or compilation. The relations mapping the rules file are formed only at the first start of the program and are being adjusted in the case of change of file modification date or creation date of load module.

## 10. SHM_db and natural language processing

Now we shall dwell upon those possibilities, which SHM_db submits for the tasks of natural language processing. For the recursive analysis of rules, described in the second section of the present paper, some changes are entered into kernel of a DBMS. The mode, distinguished from the usual one by a little bit other processing of finite states and by usage of the additional stack for preventing of loopholes at recursive search, is added. Besides, two files of rules are introduced (in a nomenclature SHM_db) with the fixed names *RuleTrue* and *RuleFalse*. At initialization of the system the files are empty, their filling is being carried out in the learning process. One more addition concerns the determining at input all possible references for sentences of the natural language to the schemes, available in the database. In other words it is possible to say, that the "understanding" process of the entered text as applied to SHM_db is equivalent to the determining the references of sentences to the schemes.

Let's consider two fragments of rules file of the demonstration application[*].

---

[*] The demonstration application can be submitted by the author for familiarizing in reply to inquiry by E-mail.

The 1-st fragment is creation and modification of semantic rules in the learning process (Fig.2).

```
QUERY1 = 0411;
SAVE = 0423;
DELETE = 0424;

                        /*** Teaching ***/

 ->  #win11(s1 q1 a1) | (key == QUERY1),
                                (s1 := "Tom read ( a book ) ."),
                                (q1 := "who read ( a book ) ?"),
                                (a1 := "Tom read ( a book ) .");
#win11(s1 q1 a1) - > #winR() | (key == SAVE), #add_rule0(q1 s1 a1);

#win11(s1 q1 a1) -> #winQ () | (key ==DELETE), #del_rule0(q1 s1 a1);

                         . . . . . .
```

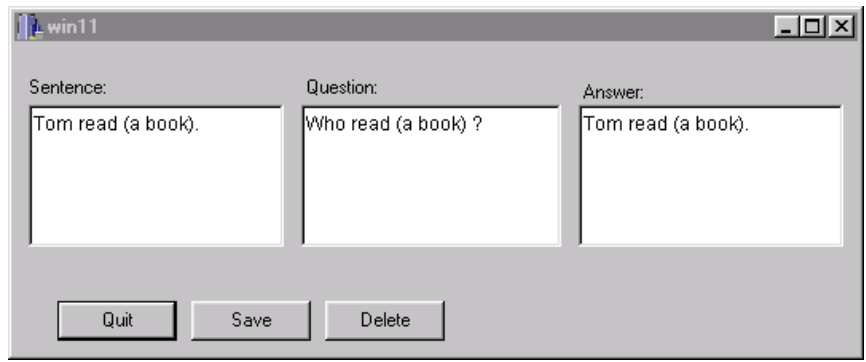

Figure 2.

Here variable s1 corresponds to the entered sentence, q1 - to the question, a1 - to the answer to the question. The value assignment for variables s1, q1, a1 is made for visualization. The parentheses select the steady collocations at training; it gives more large generalizations at modification of the schemes. The subroutine/function #add_rule0(q1 s1 a1) realizes the search of similar rules, creation or modification of the schemes of rules and addition of paradigm values. This subroutine creates also direct references (references of belonging) of created schemes to the rules file *RuleTrue*. The subroutine/function #del_rule0(q1 s1 a1) deletes the relations corresponding to the rule entered, and at impossibility of their deleting from the generalized schemes of rules, sets the given rule into the file *RuleFalse*.

Except training under the circuit *sentence-question-answer* in the demonstration application the circuits *condition-consequence* and double *condition-consequence* are realized also.

The second fragment demonstrates the example of "semantic search", in which the context search of articles from the file *Text* and the answer to the entered question according to semantic rules generated by the system in the learning process are shown (Fig.3).

```
T = (art+ {s});
QUERY4 = 0414;
SEARCH = 0427;

                    /*** Semantic search ***/

-> #win4a(q) | (key == QUERY4), (q := "who read a book ?");

#win4a (q) ->  (Flag :=0), #trans4(q s),
               (Flag ==0), #win2b ("I do not know." " ")
                                              | (key == SEARCH);

#trans4(q s), { T,} ->  (Art := art+), (Q :=q), #trans3 ({s} s1);

#trans3 ({s} s1) ->   #trans2 (Q s1 a);

#trans2 (Q s1 a)   ->  (Flag :=1), #win2b(a Art);
```

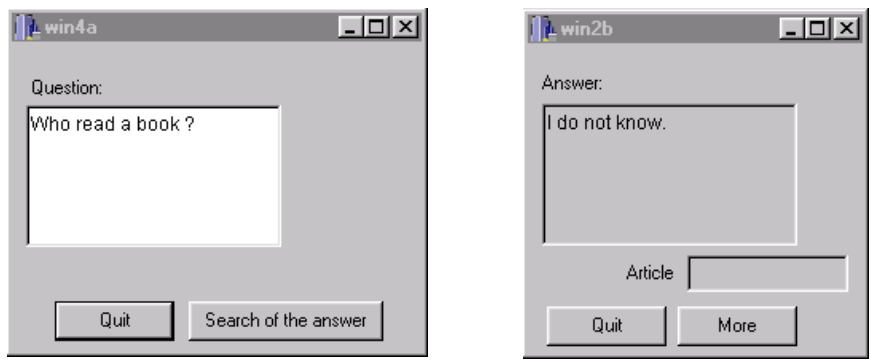

Figure 3.

`#trans4(q s)` realizes a search of sentences satisfying the condition of search, that is a search of sentences `s`, which together with a question `q` correspond to any inverse rule. And not only sentences, directly responding to the inquiry, are taken into account, but also sentences them generating.
`#trans3({s} s1)` selects the sentences immediately contained in the article retrieved, and also generates sentences being derivative from the sentences of the article and their combinations.
`#trans2 (Q s1 a)` searches the rules, satisfied by pair `Q s1` (here `Q` is the initial question, `s1` is the sentence generated by subroutine/function `trans3`) and forms the answer.

*Note*. The reverse order of variables `q` and `s` in the argument list of subroutines/functions `trans2`, `trans4` is explained by search algorithm features of the inference rules and by record of inverse rules.

The application generates not only direct answers, but also answers being derivative as for the questions (at the expense of the analysis of inverse rules) and for the answers as well (using direct rules). Assume the example, that in the database there are rules:

*(P1 is elder than P2. Who is elder than P2 ?)  ->  (P1 is elder than P2.);*

*(P2 is younger than P1.)  ->  (P1 is elder than P2.);*

*((P3 is younger than P2.)(P2 is younger than P1.) )  ->  (P3 is younger than P1.);*

Here *P1*, *P2*, *P3* are paradigms. We admit also, that article *N* contains sentences:

*Tom is younger than Bill.*

*Bill is younger than Jon.*

Then to the question:   Who *is elder than Tom?*

 two answers will be obtained:      *Bill is elder than Tom.*

*Jon is elder than Tom.*

The experiments with the program of semantic search suggest the idea, that (contrary to Chomsky ideas [2], being the theoretical base of modern compilers) a "concept" or a "semantic image" in SHM_db can be treated as some aggregate of the connected points in the space of recursive inferences. And natural language is taken as the basis of representation of information, which does not require further interpretation.

## 11.   Some problems, which require further solution

The problem of convolution of the relations. Now on the 0-th level at data input the unambiguous convolution of character sequences on component combinations always is being carried out. By its means more compact allocation of the data in base is achieved. For example, if in the database there already exists combination of characters **(in)**, the word "**input**" will be represented as the relation **((in)put).**

On the 1-st level in the semantic mode the convolution on collocations is also being realized. But the analysis on a multivalent of convolution on the 1-st level now is not being carried out. Besides, there is a non-realized possibility of creation of nested paradigms at rules modification.

The problem of anaphoric similarity requires additional study. It is closely connected with the problems of the relations' convolution and gives additional information for paradigms generalization. It results, in the long run, in increasing the speed of training (more precisely, it decreases the amount of the initial rules presented as samples).

The most essential effect at training is achieved by the generalization of paradigms. This process cannot go completely automatically and requires obtaining additional information (under the initiative of the system). The generalization of paradigms in the dialogue for the semantic mode SHM_db is now on the stage of debugging.

Taking into account, that the overall performance of the system as a whole directly depends on the speed of search of the derivative relations, the studies of algorithms of ordering the relations on their usage probability were carried out. SHM_db allows doing it by collection of statistical information in an additional shared segment with subsequent on-pair reordering of the relations by means of the service defragmentation procedure. The loss of productivity in the mode of collection of statistical information practically was not observed.

## 12. Conclusion

1. The principle of construction of DBMS with the usage of the File Mapping mechanism of the operating system has shown its ability for work. With the help of SHM_db some systems concerning the tasks of management automation and CAD were realized. Limitations on the data size concerned with 4 Gbytes address space of the 32 bit processors virtual storage will be removed a bit later if to take into consideration the tendency of development of computer equipment in the direction of microprocessors digit capacity increase.
2. DBMS, using a non-byte data representation with fixing the relations on the space of virtual addresses, despite seeming wastefulness concerning both the size of data and time of processing, is quite comparable to commercial implementations of DBMS, including relational ones. For information: the dictionary of 10000 Russian words, including their possible forms, occupies about 10 Mbytes of the virtual storage.
3. The language for record of rules, considered in the paper, includes the peculiarities permitting to realize algorithms, based on recursive inference, in the combination with procedural sequences.
4. Using the same structure of the relations for the record of both the language constructions and the data, the language expresses inquiries to the database within the framework of the syntax.
5. The features of structuring of the information in SHM_db allow adequately representing semantic structures, characteristic for natural languages. The application, described in the paper, demonstrates not only the search on the contents, but also the opportunity of reasoning simulation.
6. SHM_db could be used as a base for realization of research in the field of processing the natural language, similarly to the system GATE [4], though in other direction.

## References

1. Androutsopoulos I., Ritchie G., Thanisch P. Time, Tense and Aspect in Natural Language Database Interfaces. *Natural Language Engineering*, 4(3), pp. 229-276, Sept. 1998, Cambridge Univ. Press. Available in the Computation and Language archive under cmp-lg/9803002.
2. Higman B. *A comparative study of programming languages*. Mir, Moscow, 1974.
3. Cardie C. Empirical Methods in Information Extraction, *AI Magazine* 18 (4), pp. 65-79, 1997.
4. Cunningham H., Gaizauskas R., Humphreys K., Wilks Y. Experience with a Language Engineering Architecture: Three Years of GATE. In *Proceedings of the AISB'99 Workshop on Reference Architectures and Data Standards for NLP*, The Society for the Study of Artificial Intelligence and Simulation of Behaviour, Edinburgh, U.K. Apr, 1999.

5. Lashmanov A.V. Deductive databases: language aspect. *Databases and hypertext systems*, 9, pp. 1-33, ELDOC, Moscow, 1991.
6. Nilsson U. and Maluszynski J. *Logic, Programming and Prolog* (2ED), 1995. Available from URL = http://archive.comlab.ox.ac.uk/logic-prog.html.
7. *R++ User Manual*, AT&T Bell Laboratories, 1995. Available from URL = http://www.research.att.com/sw/tools/r++/publications.html.
8. Tang L.R. and Mooney R.J. Automated Construction of Database Interfaces: Integrating Statistical and Relational Learning for Semantic Parsing, *Proceedings of the Joint SIGDAT Conference on Empirical Methods in Natural Language Processing and Very Large Corpora (EMNLP/VLC-2000)*, pp. 133 - 141, Hong Kong, October, 2000.
9. Turing A.M. Computing Machinery and Intelligence. *The Journal of the Mind Association*, Oxford University Press, vol. LIX, 236, pp. 433-460, 1950.
10. Zelle J.M. and Mooney R.J. Learning to Parse Database Queries Using Inductive Logic Programming, *Proceedings of the 14th National Conference on Artificial Intelligence*, pp. 1050-1055, Portland, OR, August 1996. AAAI Press/MIT Press.